\documentclass[lettersize,journal]{IEEEtran}
\usepackage{amsmath,amsfonts}
\usepackage{algorithmic}
\usepackage{algorithm}
\usepackage{array}
\usepackage{textcomp}
\usepackage{stfloats}
\usepackage{url}
\usepackage{verbatim}
\usepackage{graphicx}
\usepackage{cite}
\usepackage{subcaption}
\usepackage{comment}
\usepackage{svg}
\usepackage{times}
\usepackage{epsfig}
\usepackage{amssymb}
\usepackage{wrapfig}
\usepackage{makecell}
\usepackage{xcolor}
\usepackage[subdued,italic]{mathastext}
\usepackage{rotating}
\usepackage{multirow}
\usepackage{tabularx}
\usepackage{svg}
\usepackage{threeparttable}
\usepackage[normalem]{ulem}

\begin{document}

\title{Fine-grained Classification of Solder Joints with $\alpha$-skew Jensen-Shannon Divergence}

\author{Furkan~Ulger,~Seniha~Esen~Yuksel,~\IEEEmembership{Senior Member,~IEEE}, Atila~Yilmaz,~and~Dincer~Gokcen,~\IEEEmembership{Member,~IEEE}

	\thanks{F. Ulger, S.E. Yuksel, A. Yilmaz, and D. Gokcen are all with the Department of Electrical and Electronics Engineering, Hacettepe University, Ankara, 06800, Turkey. F. Ulger is also with Aselsan Inc. 06830, Ankara, Turkey. \newline
	e-mail: furkan.ulger@stu.ee.hacettepe.edu.tr,  eyuksel@ee.hacettepe.edu.tr,\\ ayilmaz@ee.hacettepe.edu.tr, dgokcen@ee.hacettepe.edu.tr}}

\markboth{Submitted to IEEE Transactions on Components, Packaging and Manufacturing Technology}%
{Shell \MakeLowercase{\textit{et al.}}: A Sample Article Using IEEEtran.cls for IEEE Journals}

\IEEEpubid{0000--0000/00\$00.00~\copyright~2021 IEEE}

\maketitle

\begin{abstract}
Solder joint inspection (SJI) is a critical process in the production of printed circuit boards (PCB). Detection of solder errors during SJI is quite challenging as the solder joints have very small sizes and can take various shapes. 
In this study, we first show that solders have low feature diversity, and that the SJI can be carried out as a fine-grained image classification task which focuses on hard-to-distinguish object classes. To improve the fine-grained classification accuracy, penalizing confident model predictions by maximizing entropy was found useful in the literature. Inline with this information, we propose using the $\alpha$-skew Jensen-Shannon divergence ($\alpha$-JS) for penalizing the confidence in model predictions. 
We compare the $\alpha$-JS regularization with both existing entropy-regularization based methods and the methods based on attention mechanism, segmentation techniques, transformer models, and specific loss functions for fine-grained image classification tasks. We show that the proposed approach achieves the highest F1-score and competitive accuracy for different models in the fine-grained solder joint classification task. Finally, we visualize the activation maps and show that with entropy-regularization, more precise class-discriminative regions are localized, which are also more resilient to noise. Code will be made available here upon acceptance.
\end{abstract}


\section{Introduction}

\label{sec:intro}

\begin{figure*}
     \centering
     \begin{subfigure}[b]{0.3\textwidth}
         \centering
        \includegraphics[width=0.3\textwidth]{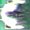}
        \hspace{.05cm}
        \includegraphics[width=0.3\textwidth]{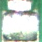}
        \hspace{.05cm}
        \includegraphics[width=0.3\textwidth]{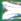}
        
        \vspace{0.30cm}
        \includegraphics[width=0.3\textwidth]{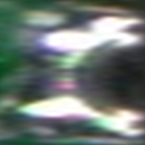}
        \hspace{.05cm}
        \includegraphics[height=1.65cm,width=1.65cm]{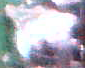}
        \hspace{.05cm}
        \includegraphics[height=1.6cm,width=1.6cm]{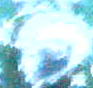}
         \caption{}
         \label{fig:y equals x}
    \end{subfigure}
    \hfill
         \begin{subfigure}[b]{0.3\textwidth}
         \centering
        \includegraphics[width=\textwidth, height= 3.5cm]{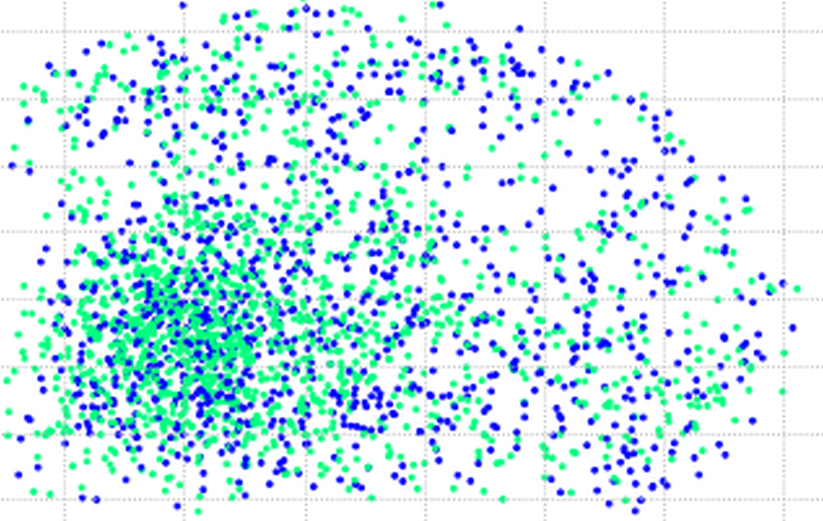}
         \caption{}
         \label{fig:five over x}
     \end{subfigure}
          \hfill
    \begin{subfigure}[b]{0.3\textwidth}
        \centering
        \includegraphics[width=\textwidth, height= 3.5cm]{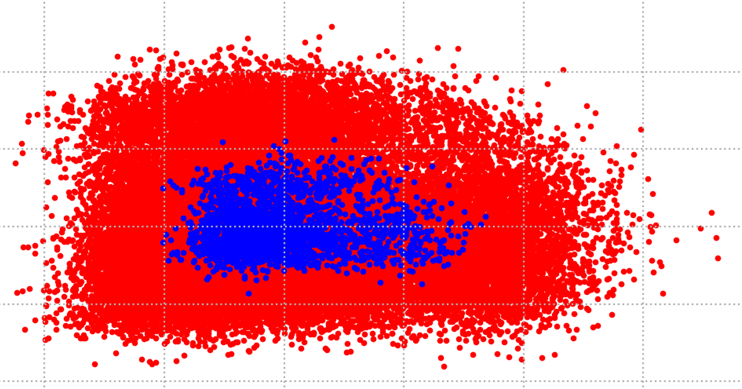}
        \caption{}
        \label{fig:three sin x}
     \end{subfigure}
    \caption{(a) Samples from the solder joint dataset: normal solders are at the top, defective solders are at the bottom. (b) PCA projection of normal and defective solder joints' features are given in blue and green respectively; showing that the inter-class variance among the normal and defective solder classes is low, but the intra-class variance is high. (c) PCA projection of the solder dataset on the last fully connected layer of VGG16 is shown in blue, ImageNet in red; showing that the solder image features' diversity is much less than ImageNet. }
    \label{fig:pca}
\end{figure*}

In the electronics manufacturing industry, most of the errors during PCB production are caused by solder joint errors; therefore, inspection of solder joints is an important procedure in the electronic manufacturing industry. Accurate detection of solder errors reduces manufacturing costs and ensures production reliability. Solder joints are generally very small and form different shapes on PCBs. The main difficulty in visual inspection of solder joints is that there is generally a minor visual difference between defective and defect-free (normal) solders \cite{moganti1996automatic}. Even a small change in the form of the solder joint can pose a fatal error in PCBs. 
Considering these, we propose approaching the problem as fine-grained image classification (FGIC), where inter-class variations are small and intra-class variations are large, which is the opposite of generic visual classification. 
Several examples of normal and defective solders are shown in Fig.~\ref{fig:pca} (a). We compare the diversity of the solder joint features to the generic images of ImageNet \cite{imagenet2009} in Fig.~\ref{fig:pca} (b).  
For visualization, we sketch the top two principle components of the output features on the last fully connected layer of the deep Convolutional Neural Network (CNN) architecture VGG16 \cite{simonyan2014very} for the solder dataset and ImageNet; and observe that  the solder image features' diversity is much less than ImageNet. Blue and green samples in Fig.~\ref{fig:pca} (c) represent normal and defective solder joints' features respectively. Feature diversity of the samples across the classes is similar as they show a low inter-class variance, whereas the features are diverse for the same class since they present a high intra-class variance. These observations on the dataset show that SJI is a fine-grained image classification problem.

As opposed to generic image classification, samples that belong to different classes can be visually very similar in FGIC tasks. 
In this case, it is reasonable to penalize the classifiers that are too confident in their  predictions, which in turn reduces the specificity of features and improves generalization by discouraging low entropy \cite{wei2021fine}. One way to achieve FGIC is through entropy-regularization. 
Entropy-regularization based methods such as label smoothing \cite{szegedy2016rethinking} and maximum entropy learning \cite{dubey2018maximum} penalize model predictions with high confidence. 
An alternative entropy-regularization technique is the $\alpha$-JS \cite{nielsen2010family, nielsen2011burbea}. In fact, maximum entropy learning and label smoothing are two extremes of $\alpha$-JS \cite{meister2020generalized}. 


In this study, we show that solder joints are actually fine-grained, and that their inspection can be 
treated as a FGIC problem. Then,  we propose using $\alpha$-JS as an entropy-regularization method that improves inspection accuracy over the other entropy-regularization methods. Additionally, we compare the proposed $\alpha$-JS regularization method to recent FGIC methods that are based on different approaches  such as employing segmentation techniques, attention mechanism, transformer models, and designing specific loss functions. We show that regularizing with $\alpha$-JS achieves the highest F1-score and competitive accuracy levels for different models in fine-grained classification of solder joints. 
\IEEEpubidadjcol

Our contributions are as follows: ($1$) We propose using $\alpha$-JS as an entropy regularization in fine-grained classification tasks for the first time. ($2)$ We illustrate that SJI is a FGIC task. ($3$) We show that entropy regularization with $\alpha$-JS improves classification accuracy over the existing methods across different models on fine-grained solder joints. ($4$) We demonstrate that regularizing entropy with $\alpha$-JS can focus on more distinctive parts in the image, and this improves the classification accuracy.

We employ gradient-weighted class activation mapping (Grad-CAM) \cite{selvaraju2017grad}, an activation map visualization method, to see the effect of regularizing entropy on the activation maps 
to classify the solder joints. We find that localized regions of the image used in classification are more reasonable for a model trained with entropy-regularization and less affected by background noise. Although the proposed regularization method is tested for the fine-grained solder joints dataset, it holds potential for other FGIC datasets as well.

\section{Related Work}
\label{sec:format}

\textbf{Solder joint inspection} is a long standing problem in the literature \cite{moganti1996automatic, abd2020review}. Proposed approaches can be divided into $3$ groups: reference-based, traditional feature-based, and deep learning based. Reference-based methods use a defect-free (template) version of the defective PCB to compare. This comparison can be made in several ways, including pixel-wise subtraction to find the dissimilarities \cite{west1984system, wu1996automated}; template matching via normalized cross correlation to find the similarities \cite{annaby2019improved, crispin2007automated}; frequency domain analysis \cite{tsai2018defect}; and template image modelling using feature manifolds \cite{jiang2012color}, Principle Component Analysis \cite{cai2017ic}, and gray values statistics \cite{xie2009high}. The generalization of referential methods rely on the number of templates of the defective image, and this restricts the practical applications.



In the traditional feature extraction based methods, color contours \cite{capson1988tiered, wu2013classification}, geometric \cite{acciani2006application, wu2014inspection}, wavelet \cite{acciani2006application, mar2011design}, and gabor \cite{mar2011design} features have been used followed by a classifier. In order to extract these features, solder joints are required to be segmented or localized from PCBs. 
For segmentation, colour transformation and thresholding \cite{jiang2007machine, mar2011design}, as well as colour distribution properties \cite{zeng2011algorithm} and Gaussian Mixture Models \cite{zeng2011automated} under a special illumination condition
are used for segmentation. Then, to extract features, the localized solder joints are divided into subregions \cite{hongwei2011solder, wu2011feature},  the selection of which can also be optimized as in \cite{song2019smt}.


After the introduction of CNN, deep learning methods outperformed the traditional methods in solder inspection, despite a need for a lot of data. 
Although defectless (normal) solder joints are easy to obtain, defective solder joints are quite rare. To alleviate this problem, some studies modelled the normal solder data distribution with generative models \cite{li2021ic, Ulger2021}.
When enough data is available, Cai et al. \cite{cai2018smt} obtained very high accuracy with CNNs in solder joint classification under special illumination conditions. Li et al. \cite{li2020automatic} utilized an ensemble of Faster R-CNN and YOLO object detectors to both localize a solder from PCB and to classify it. YOLO detector is also employed on thermal PCB images to detect electronic component placement and solder errors in \cite{jeon2022contactless}. A lightweight custom object detector is designed by fusing convolutional layers for defect localization in \cite{wu2022pcbnet}. The two surveys \cite{abd2020review, moganti1996automatic} also provide a comprehensive review on SJI methods.


\textbf{Fine-grained image classification}. In this study, we further analyse the solder joints in another frame and show that they have less feature diversity, and form fine-grained images. By regularizing the entropy in model training, we improve the classification accuracy. The main challenge in FGIC is that classes are visually quite similar and intra-class variation is large in contrast to generic classification. In the literature, mainly (i) localization-classification subnetworks and (ii) end-to-end feature encoding are applied to this problem \cite{wei2021fine}. 
\textit{Localization-classification subnetworks}
 aim to localize image parts that are discriminative for FGIC. 
In \cite{zhang2019learning} expert subnetworks that learn features with attention mechanism are employed  to make diverse predictions. In \cite{ji2020attention}, attention convolutional binary neural tree architecture (ACNet) is introduced with an attention transformer module in a tree structure to focus on different image regions. In \cite{behera2021context}, a context-aware attentional pooling module is used to capture subtle changes in images. 
In \cite{rao2021counterfactual}, causal inference is used  as an attention mechanism.
\textit{End-to-end feature encoding} tries to learn a better representation to capture subtle visual differences between classes. This can be achieved by designing specific loss functions to make the model less confident about its predictions \cite{szegedy2016rethinking, pereyra2017regularizing}. During model training, cross-entropy between the model and target distribution is minimized, which corresponds to maximizing likelihood of a label; however this can cause over-fitting. Several methods are proposed to increase model generalization. Szegedy et al. \cite{szegedy2016rethinking} propose label smoothing regularization that penalizes the deviation of predicted label distribution from the uniform distribution \cite{szegedy2016rethinking}. Dubey et al. \cite{dubey2018maximum} show that penalizing confident output distributions \cite{pereyra2017regularizing} improves accuracy in FGIC. Mukhoti et al. \cite{mukhoti2020calibrating} show that focal loss \cite{lin2017focal} can increase the entropy of model predictions to prevent overconfidence. 
Meister et al. \cite{meister2020generalized} derived an alternative entropy-regularization method from $\alpha$-JS, called as generalized entropy-regularization, and showed its success in language generation tasks.
Designing loss functions is not limited to entropy regularization. 
For instance, the mutual channel loss in \cite{chang2020devil} forces individual feature channels belonging to the same class to be discriminative. 



Recently, different approaches are also proposed to tackle the FGIC task. Du et al. \cite{du2020fine} used a jigsaw puzzle generator to encourage the network learn at different visual granularities. Luo et al. proposed Cross-X learning \cite{luo2019cross} to exploit relationship between features from different images and network layers. 
He et al. \cite{he2022transfg} added a part selection module to Transformer models for FGIC tasks, which integrates attention weights from all layers to capture the most class distinctive regions. In the results section, we compare our approach to both the entropy-regularization and these alternative methods.

The goal in regularizing the entropy is to get the most unbiased representation of the model which is achieved by the probability distribution with maximum entropy \cite{jaynes1957information}. Several methods have been proposed to increase generalization of a classifier by maximizing entropy. The classifier is maximally confused when the probability distribution over the model predictions is uniform, where the classifier assigns the same probability in predicting between all the classes. Three alternatives to entropy-regularization are given below.

\textit{Focal loss} \cite{lin2017focal} adds a weighted term to the cross-entropy loss to put more weight on the misclassified examples. Additionally, it ensures entropy of the predicted probability distribution to be large. This leads to less confident predictions and better generalization. Focal loss $\mathcal{L}_{f}(\theta)$ is defined in Eq.~\ref{eq:focal} where $\alpha  \in  [0,1]$, and $\gamma$ is a scalar value. $y_i \in \{0, 1\}$ represents the ground-truth class, and $p$ is the  probability of the sample to belong to class $y_{i}=1$ as given in Eq.~\ref{eq:pt}.

\begin{equation}
    \mathcal{L}_{f}(\theta) =  -\alpha_{t} (1-p_{t})^\gamma \mathrm{ln}(p_{t})
    \label{eq:focal}
\end{equation}

\begin{equation}
        p_{t}= 
    \begin{cases}
        p,& \text{if } y_{i}=1\\
        1-p, & \text{otherwise}
    \end{cases}
    \label{eq:pt}
\end{equation}


\textit{Label smoothing regularization} \cite{szegedy2016rethinking} replaces 
target label distribution by a smoothed distribution to make the model less confident about its predictions. It encourages 
model distribution $p_{\theta}$\footnotemark[1] to be close to uniform distribution $u$ to maximize entropy. This is achieved by minimizing cross-entropy $\mathrm{H}(u,p_\theta)$ as given in Eq.~\ref{eq:lsr}, where $\varepsilon$ is a 
smoothing term, $\varepsilon \in [0,1]$. 

\footnotetext[1]{Conditional probability distribution $p_\theta(\mathbf{y}|\mathbf{X})$ is represented by $p_\theta$ for brevity.}

\begin{equation}
    \mathcal{L}_{s}(\theta) =  (1-\varepsilon)\mathrm{H}(\mathbf{y}, p_{\theta})
    + \varepsilon \mathrm{H}(u, p_{\theta})
    \label{eq:lsr}
\end{equation}


\textit{Maximum entropy learning} \cite{dubey2018maximum} minimizes the Kullback-Leibler ($\mathrm{D_{KL}}$) divergence between the model distribution $p_{\theta}$ and the true label distribution $\mathbf{y}$.
A confidence penalty term is added to maximize the entropy of model predictions $\mathrm{H}(p_\theta)$ and make it more uniform as defined in Eq. \ref{eq:maxEnt}, where $\beta$ is a term for strength of the penalty.

\begin{equation}
    \mathcal{L}_{m}(\theta) = \mathrm{D_{KL}}(\mathbf{y}||p_{\theta}) - \beta \mathrm{H}(p_{\theta}) 
    \label{eq:maxEnt}
\end{equation}

\section{METHODS}

\begin{figure}
    \centering
    \includegraphics[width=0.48\textwidth]{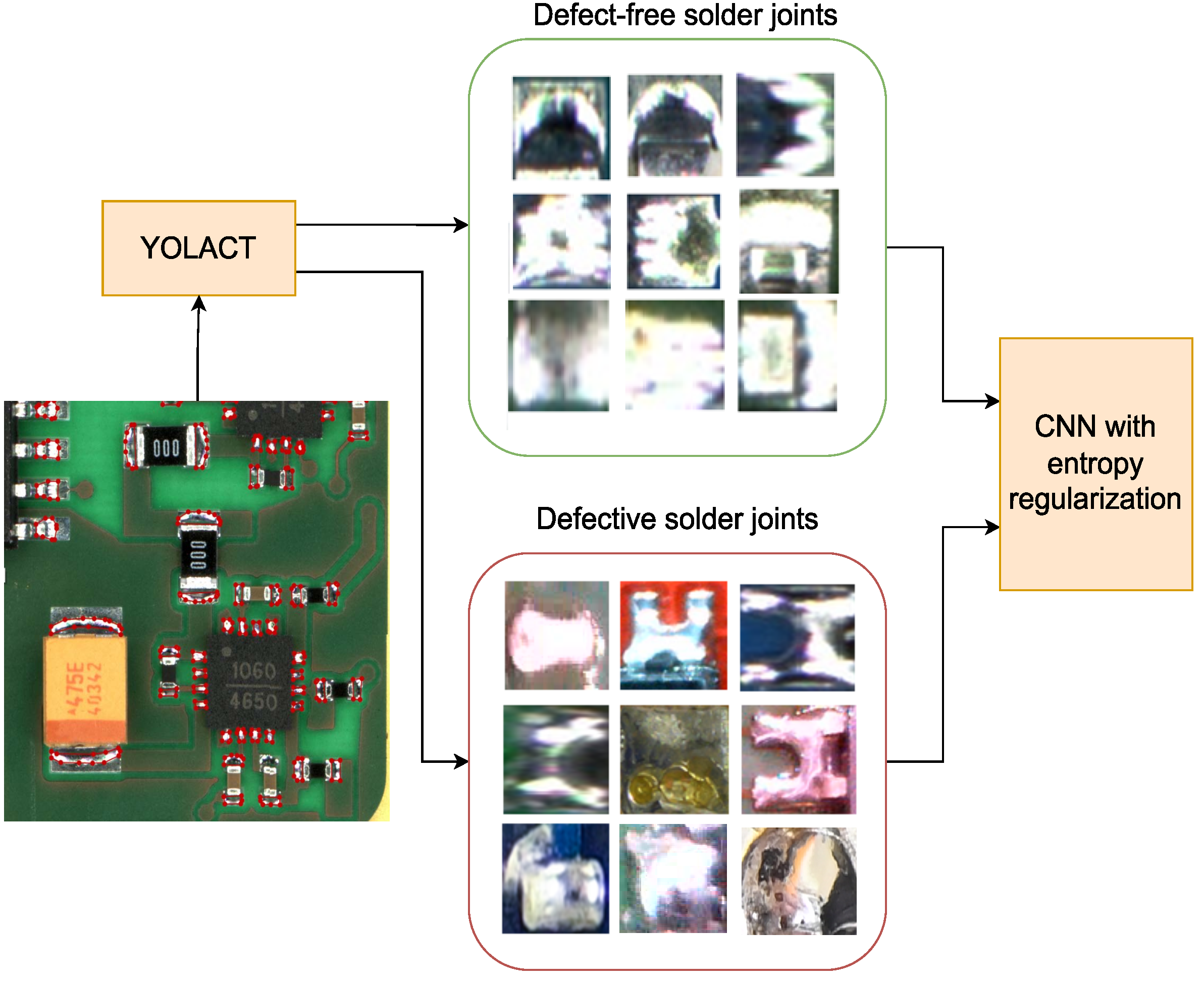}
    \vspace{0.5cm}
    \caption{Proposed model pipeline. Both defect-free (normal) and defective solder joints are segmented with YOLACT. Finally, the CNN with entropy regularization is trained with the solder joint classification.}
    \label{fig:pipeline}
\end{figure}

Our proposed pipeline includes segmentation and classsification of individual solder joints on PCBs as shown in Fig.~\ref{fig:pipeline}. Solder joints are segmented with You Only Look at Coefficients (YOLACT) \cite{bolya2019yolact}, an instance segmentation method. Normal and defective solder joints are classified with a CNN with entropy regularization.

\subsubsection{Segmentation: YOLACT}
\label{sec:segmentation}

YOLACT mainly employs a Feature Pyramid Network (FPN) as a backbone to extract features. The feature maps are used in two branches to produce prototype masks and bounding boxes with associated labels. The solder joints are segmented from $7$ high resolution PCB images that are divided into tiles for easier processing. Initially, more than $1000$ normal solder joints and $250$ defective samples are labelled on PCB tiles by annotating through Labelme \cite{labelme} in model training. Both normal and defective solder joints are segmented with YOLACT. Only imprecise segmentation results are corrected manually to create a solder dataset for FGIC. 
The proposed pipeline is given in Fig.~\ref{fig:pipeline}. All the individual solder joints are obtained as a result of this segmentation.

\subsubsection{Classification with Entropy Regularization: $\alpha$-JS}

In supervised classification, the objective to minimize the entropy between the model distribution and distribution over the labels leads to overconfidence in assigning probabilities; and this can cause overfitting \cite{szegedy2015going}. To alleviate this problem, overconfidence is penalized by maximizing the entropy. 
A way to maximize entropy is encouraging model entropy to be as close as possible to uniform distribution, where the entropy is maximum. We propose using $\alpha$-JS as an entropy-regularization method in fine-grained classification of solder joints. $\alpha$-JS is obtained by the sum of weighted skewed Kullback-Leibler (KL) divergences by introducing a skewness parameter, $\alpha$ that determines weights for probability distributions $p$ and $q$, and scaled by $\frac{1}{\alpha(1-\alpha)}$ as given in Eq.~\ref{eq:JSD_KL}. Derivation is available in Appendix B.

\begin{multline}
    \mathrm{J}^s_{\alpha}(p||q)
    = \frac{1}{\alpha(1-\alpha)}[(1-\alpha)\mathrm{D_{KL}}(p||(1-\alpha) p+\alpha q) + \\ \alpha \mathrm{D_{KL}}(q||(1-\alpha) p+\alpha q)] 
    \label{eq:JSD_KL}
\end{multline}

The penalty term given in Eq. \ref{eq:alpha-JS} is added to encourage high entropy by minimizing the distance between the model and uniform distribution.

\begin{multline}
    \textrm{J}^s_{\alpha}(u||p_{\theta}) = \frac{1}{\alpha(1-\alpha)}[- \alpha \mathrm{H}(p_{\theta}) - (1 - \alpha)\mathrm{H}(u)+\\ \mathrm{H}((1-\alpha)u + \alpha p_{\theta})]
     \label{eq:alpha-JS}
\end{multline}

As a result, the optimized objective function to train the model is given in Eq. \ref{eq:JSDreg}. 

\begin{equation}
    \mathcal{L}_{\mathrm{J}^s_\alpha}(\theta) = \mathrm{H}(\mathbf{y}, p_{\theta}) + \beta  \mathrm{J}^s_{\alpha} (u || p_{\theta})
    \label{eq:JSDreg}
\end{equation}

In fact, using $\alpha$-JS with $\alpha \rightarrow 0$ as a regularizer is equivalent to maximum entropy learning, and label smoothing regularization with $\alpha \rightarrow 1$. The derivation is available in \cite{meister2020generalized}.

In this study, we propose using $\alpha$-JS as an entropy regularizer as given in Eq.~\ref{eq:JSDreg}. The skewness parameter ($\alpha$) in $\alpha$-JS allows us to set different entropy values for the model by giving different weights to the uniform and model distributions. 
Since we observe that higher entropy does not necessarily correspond to higher classification accuracy, which is observed with the experiments in Sec. \ref{sec:results}, by tuning the $\alpha$ parameter and testing different entropy values, we show that higher accuracy can be achieved for FGIC of solder joints across different models.

\section{Dataset and Experiments}
\label{sec:typestyle}


\textbf{Dataset:} As a result of the segmentation in Section \ref{sec:segmentation}, the solder dataset has $3292$ normal and $760$ defective solder joints. The labels are confirmed by experts. The type of solder joint errors include mostly solder bridge, excessive, and insufficient solder errors. Additionally, there are some cracked solder, voids, shifted component, corrosion and flux residue errors. About $20\%$ of the dataset that corresponds to $659$ normal and $227$ defective solder joints are set aside randomly as the test set. The rest of the dataset is used for stratified five-fold cross validation, which retains a percentage of samples for each class in the folds, due to class imbalance. The segmented solder image sizes vary from $8\times8$ to $180\times180$, therefore we trained the models with different image sizes: $28\times28$, $84\times84$, and $224\times224$. Image size of $28\times28$ decrease accuracy, and large image size require a lot of training time; hence all the samples were resized to $84 \times 84$. The data is normalized to have zero mean and unit variance prior to training.

\noindent \textbf{Experimental settings:} We experiment on different architectures such as GoogleNet \cite{szegedy2015going}, VGG16 \cite{simonyan2014very}, ResNet-18 and ResNet-50 \cite{he2016deep} with the same initial weights to evaluate the robustness of the proposed method in FGIC. All the models are trained for $5000$ to $10000$ iterations until convergence with batch size of $64$ samples. Root Mean Squared Propagation (RMSprop) with momentum and Adam optimizer \cite{kingma2014adam} is used interchangeably with an initial learning rate of $1e-4$. $\textrm{L}_2$ regularization is applied to mitigate overfitting. 
The final results are reported by the model that performed the best in the validation set with either minimum validation loss or maximum validation accuracy, then the model is tested in the test set. 
Entropy regularization strength term $\beta$ is set to $1$ for both maximum entropy learning and $\alpha$-JS for fair comparison. For label smoothing, $\varepsilon=0.1$ is selected, and focal loss is experimented with $\alpha_{t}=0.25$ and $\gamma= 2$, that is widely used in the literature. F1-score is used as an evaluation metric due to the imbalance in the dataset. 


\subsection{Experimental Results} \label{sec:results}

We compare the proposed approach with (i) entropy-regularization based models in Sec. \ref{sec:results1}, and (ii) the models that employ segmentation techniques, attention mechanism, transformer networks and new loss functions in Sec. \ref{sec:results2}.

\begin{table}[!t]
    \begin{minipage}{0.98\linewidth}
    \centering
    \begin{tabularx}{\textwidth}{|c|c||c|c|c|c|} 
     \hline
     \multicolumn{2}{|c||}{Regularization} & VGG16 & ResNet18 & ResNet50 & GoogleNet \\ [0.5ex] 
     \hline\hline
     \multicolumn{2}{|c||}{No regularization} & \textcolor{black}{96.599} & \textcolor{black}{96.035} & \textcolor{black}{96.629} & \textcolor{black}{97.297} \\ 
     \hline
     \multicolumn{2}{|c||}{Focal loss \cite{lin2017focal}} & \textcolor{black}{96.903} & \textcolor{black}{96.847} & \textcolor{black}{96.833} & \textcolor{black}{97.321} \\
     \hline
     \multicolumn{2}{|c||}{\thead{Label smoothing  \\ ($\epsilon= 0.1)$ \cite{szegedy2015going}}} & \textcolor{black}{97.285} & \textcolor{black}{96.018} & \textcolor{black}{98.206} & \textcolor{black}{96.916} \\
     \multicolumn{2}{|c||}{\thead{Label smoothing \\ ($\alpha \rightarrow 1$) \footnotemark[2] }} & \textcolor{black}{97.118} & \textcolor{black}{96.429} & \textcolor{black}{97.978} & \textcolor{black}{97.297} \\
     \hline
     \multicolumn{2}{|c||}{\thead{Max. entropy \\ ($\alpha \rightarrow 0$) \footnotemark[3] \cite{dubey2018maximum} }} & \textcolor{black}{97.297} & \textcolor{black}{96.689}
     & \textcolor{black}{97.039} & \textcolor{black}{97.528} \\
     \hline  
     \parbox[t]{3mm}{\multirow{4}{*}{\rotatebox[origin=c]{90}{$\alpha$-JS}}} &
     $\alpha= 0.1$ & \textcolor{black}{$\mathbf{98.198}$} & \textcolor{black}{96.145} & \textcolor{black}{97.309}
     & \textcolor{black}{97.788} \\
     & $\alpha= 0.5$ & \textcolor{black}{$97.321$} & \textcolor{black}{$\mathbf{97.333}$} & \textcolor{black}{$\mathbf{98.441}$} & \textcolor{black}{$\mathbf{98.214}$} \\
     & $\alpha= 0.75$ & \textcolor{black}{97.297} & \textcolor{black}{96.380} & \textcolor{black}{96.833} & \textcolor{black}{$\mathbf{98.214}$} \\
     & $\alpha= 0.9$ & \textcolor{black}{97.065}
     & \textcolor{black}{97.321} & \textcolor{black}{96.889} & \textcolor{black}{97.987} \\ [1.25ex] 
     \hline
    \end{tabularx}
    \captionof{table}{F1-score (\%) of different entropy-regularization methods on solder test set across different model architectures. The highest accuracy for each model is shown in bold 
    } 
    \label{tbl:f1_results}
    \end{minipage}
\end{table}%


\begin{table}[!t]
    \centering
    \begin{tabular}{|c|c|c|c|}
        \hline
        Method & Backbone & Accuracy & F1-score  \\
        \hline \hline
        ACNet \cite{ji2020attention} & VGG16 & $96.720$ & $93.570$ \\ \hline
        AP-CNN \cite{ding2021ap} & VGG16 & $95.559$ & $91.1765$ \\ \hline
        CAP \cite{behera2021context} & VGG16 & $98.068$ & $96.754$
        \\ \hline
        MC-Loss \cite{chang2020devil}  & VGG16 & $86.636$ & $64.880$ \\
        \hline
        DFL \cite{wang2018learning} & VGG16 & $\underline{98.867}$ & $\underline{97.758}$ \\ \hline
        Cross-X \cite{luo2019cross} & ResNet50 & $98.980$ & $97.996$
        \\ \hline
        CAL \cite{rao2021counterfactual} & ResNet50 & $\mathbf{99.450}$ & $\mathbf{99.107}$ \\ \hline
        WS-DAN \cite{hu2019see} & ResNet50 & $98.188$ & $96.491$ \\ \hline
        PMG \cite{du2020fine}  & ResNet50 & $98.402$ & $95.873$
        \\ \hline
        MGE-CNN \cite{zhang2019learning} & ResNet50 & $98.410$ & $96.804$ \\ \hline
        TransFG \cite{he2022transfg} & ViT B/16 & $98.075$ & $96.180$ \\
        \hline \hline
        \multirow{2}{*}{$\alpha$-JS} & VGG16 & $\mathbf{99.094}$ & $\mathbf{98.198}$ \\  
         & ResNet50 & $\underline{99.207}$ & $\underline{98.441}$ \\
        \hline
    \end{tabular}
    \captionof{table}{Comparison with other FGIC approaches based on various different techniques on the solder dataset. The highest accuracy and F1-score is given in bold, and the second is underlined.}
    \label{tbl:other_vs_jsd}
\end{table}

\footnotetext[2]{Equivalent to regularizing with $\textrm{lim}_{\alpha \rightarrow 1}\textrm{J}_\alpha(u||p_\theta)$}

\footnotetext[3]{Equivalent to regularizing with $\textrm{lim}_{\alpha \rightarrow 0}\textrm{J}_\alpha(u||p_\theta)$}

\subsubsection{Comparison with Entropy-Regularization Methods} \label{sec:results1}

The results of entropy-regularization based models is given in Table \ref{tbl:f1_results}. Here, we compare the $\alpha$-JS regularization across different architectures to entropy-regularization-based methods such as label smoothing, focal loss, and maximum entropy learning as well as the models trained without regularization. These methods are chosen since they are designed to maximize the entropy for FGIC through a new loss function as opposed to changing the model architecture or employing alternative mechanisms.
The highest F1-score is achieved with $\alpha$-JS and $\alpha= 0.1$ for VGG16, $\alpha= 0.5$ for ResNet18, and ResNet50, $\alpha= 0.5$ and $\alpha= 0.75$ for GoogleNet. An improvement in accuracy by approximately $1.6\%$ for VGG16, $1.3\%$ for ResNet18, $1.8\%$ for ResNet50, and $0.9\%$ for GoogleNet is achieved over the model trained without regularization. 
The results show that entropy-regularization based methods are very effective in classifying solder joints, especially the $\alpha$-JS, which outperformed the other regularization methods.

\begin{figure}
  \centering
  \includegraphics[width=0.47\textwidth]{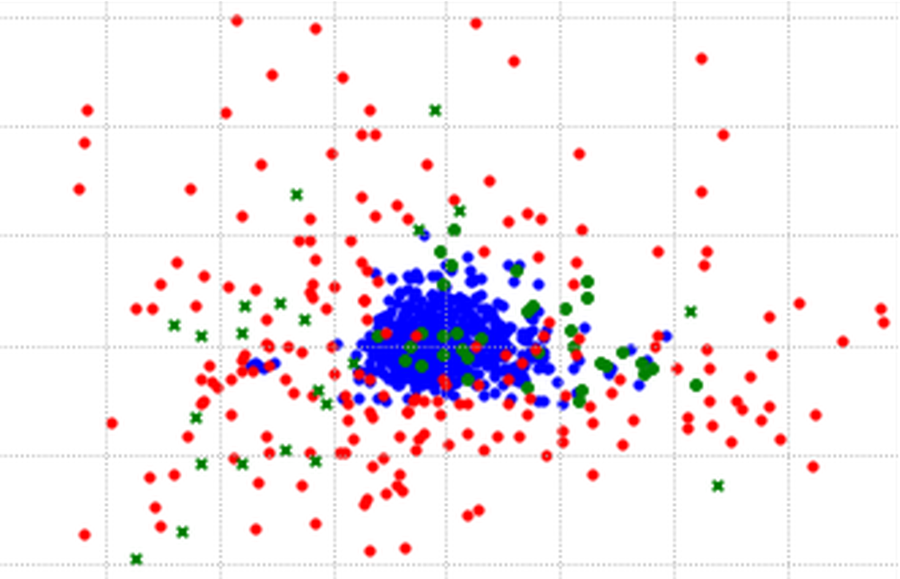} 
  \vspace*{.01pt}
  \captionof{figure}{Principle components of the test set on output features of the average pooling layer of ResNet50. Normal and defective solder joints are shown in blue and red respectively. The green points represent the normal and defective solder joints that are misclassified by the model trained without regularization, but are correctly classified when $\alpha$-JS is used as the entropy-regularization.}
  \label{fig:noreg_to_maxEnt}
\end{figure}


Effect of entropy-regularization is visualized in Fig. \ref{fig:noreg_to_maxEnt}. Note that there were many misclassifications in the vicinity of normal samples, and among the scattered defective samples as shown in green. The green circles represent the normal samples that are misclassified by the model trained without regularization (false negatives), classified correctly by $\alpha$-JS (true positives). The green crosses stand for the correctly classified defective solder samples (true negatives) by $\alpha$-JS that were misclassified otherwise (false positives).

\begin{figure}[h]
    \centering
    \includegraphics[width=\linewidth]{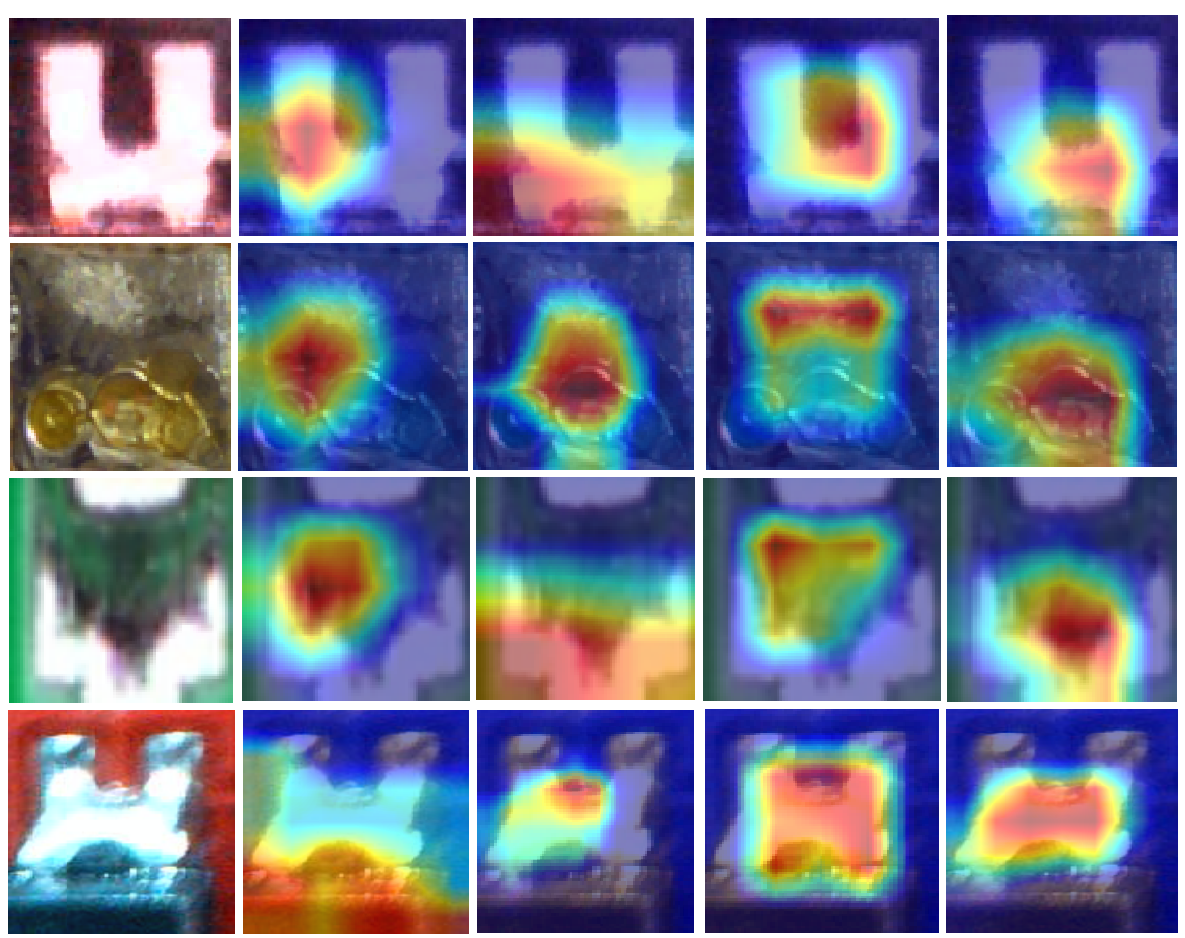} \\
    \hspace{-7.1cm} \textbf{(a)} \hspace{1.1cm} \textbf{(b)} \hspace{1.1cm} \textbf{(c)}
    \hspace{1.1cm} \textbf{(d)}
    \hspace{1.1cm} \textbf{(e)}
    \hspace{-7.2cm}
    \vspace{.15cm} 
    \caption{
    Localized class-discriminative regions by Grad-CAM. RGB test images (a). Activation maps on VGG16 trained without any regularization (b), DFL \cite{wang2018learning} (c), PMG \cite{du2020fine} on ResNet50 (d), and with $\alpha$-JS (e). The heatmap shows which part of the image that the model focuses on to make decision. The intensity increases from blue to red. The CAM results overlap the most with soldered regions for the model regularized with $\alpha$-JS.}
    \label{fig:gradCAM}
\end{figure}

In order to interpret the effect of entropy-regularization and model predictions in images, activation maps of each test image are visualized through Grad-CAM. We used the implementation in \cite{jacobgilpytorchcam} for Grad-CAM visualization. The last convolutional layer of each model are visualized to see which part of the image is used in making decision.
In Fig. \ref{fig:gradCAM}, (a) test images, (b) activation maps on VGG16 model trained without any regularization, (c) DFL \cite{wang2018learning}, (d) PMG \cite{du2020fine} on ResNet50, and (e) with $\alpha$-JS are shown. The model regularized with $\alpha$-JS yields more precise class-discriminative regions (solder joints/errors) compared to the other approaches. Shorted solder in the first and last row of Fig.~\ref{fig:gradCAM}, flux residues in the second row, and passive electronic component solder joint in the third row are localized more accurately with $\alpha$-JS.  
Additionally, it is not affected by background noise such as PCB background and non-soldered regions as much as the other models. 


\subsubsection{Comparison with the Other Approaches} \label{sec:results2}

Comparison with the other state-of-the-art approaches are presented in Table \ref{tbl:other_vs_jsd}. 
We experimented with not only the recent state-of-the-art methods based on designing FGIC-specific loss functions \cite{chang2020devil} similar to our approach, but also experimented with employing segmentation networks \cite{wang2018learning, behera2021context}, attention mechanism \cite{hu2019see, ji2020attention, ding2021ap, rao2021counterfactual}, and others \cite{zhang2019learning, du2020fine, luo2019cross, he2022transfg}. 
All the models are trained  until convergence ($5000$ iterations) from scratch, and tuned by changing their learning rate and optimizer. The best results are reported on the test set at the end of training.
The proposed approach achieves competitive results on the solder joint dataset without changing the model architecture, using attention mechanism or employing segmentation techniques. The models are compared both with respect to accuracy and F1-score. Accuracy is calculated by the ratio of correct predictions to the number of test samples. For VGG16, $\alpha$-JS leads to $0.22\%$ and $0.44\%$ improvement over the closest method \cite{wang2018learning} in terms of accuracy and F1-score respectively. The proposed regularization with $\alpha$-JS is competitive for ResNet50, closely following  \cite{rao2021counterfactual} with $0.25\%$ and $0.65\%$ in terms of accuracy and F1-score respectively at the expense of extra computational cost for the attention module. 

\begin{figure*}[t!]
    \centering
    \begin{subfigure}[t]{0.5\textwidth}
        \centering
        \includegraphics[height=1.25in]{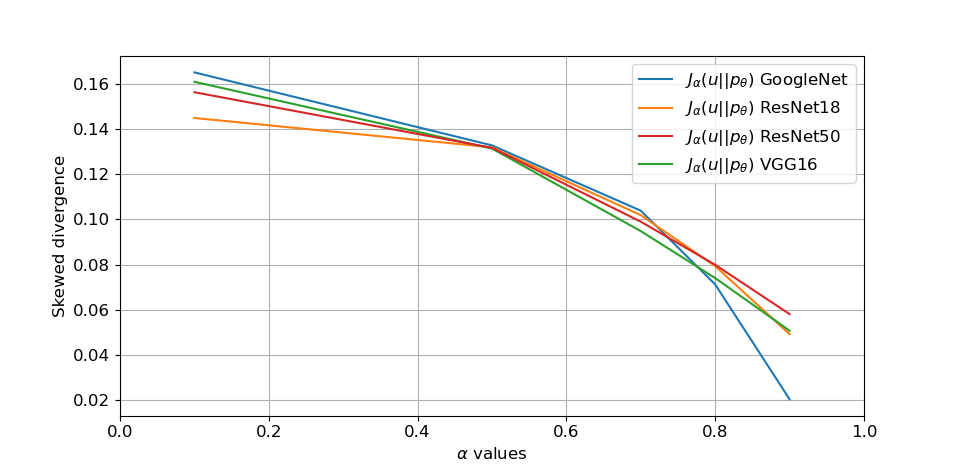}
        \caption{}
    \end{subfigure}%
    ~ 
    \begin{subfigure}[t]{0.5\textwidth}
        \centering
        \includegraphics[height=1.25in]{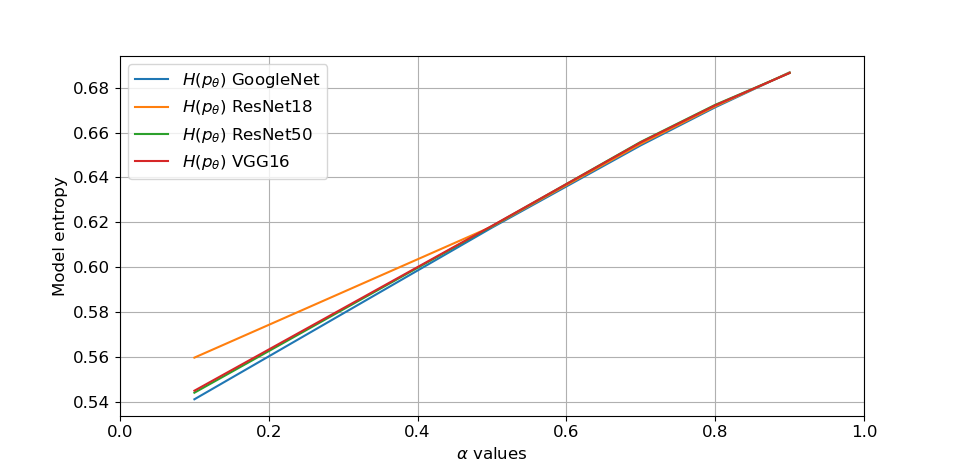}        \caption{}
    \end{subfigure}
    \vspace{0.15cm}
    \caption{$\alpha$-JS $\mathrm{J}^s_\alpha(u||p_\theta)$ (a) and model entropy $\mathrm{H}(p_\theta)$ (b) as a function of $\alpha$ for VGG16, ResNet18, ResNet50, and GoogleNet architectures. Model entropy and $\alpha$-JS are calculated for $\alpha$ = [0.1, 0.5, 0.75, 0.9].}
    \label{fig:H_J_wrt_alpha}
\end{figure*}

We investigate the effect of the skewness parameter ($\alpha$) on the $\alpha$-JS $\mathrm{J}^s_\alpha(u||p_\theta)$ for discrete binary distributions. $\alpha$-JS and entropy values as a function of $\alpha$ for VGG16, ResNet18, ResNet50, and GoogleNet are shown in Fig. \ref{fig:H_J_wrt_alpha} (a) and (b) respectively. By calculating the divergence as a function of $\alpha$, we observe that the increase in $\alpha$ resulted in lower divergence between uniform distribution $u$ and model distribution $p_\theta(\mathbf{y}|\mathbf{X})$. This monotonic relation shows that the model distribution becomes closer to the uniform distribution with higher $\alpha$ values. Taking this into consideration, getting closer to the uniform distribution yields higher model entropy $\mathrm{H}(p_\theta)$ 
as expected. The divergence of the models are calculated on the test set.

\begin{figure}
    \centering
    \includegraphics[width=8.0cm]{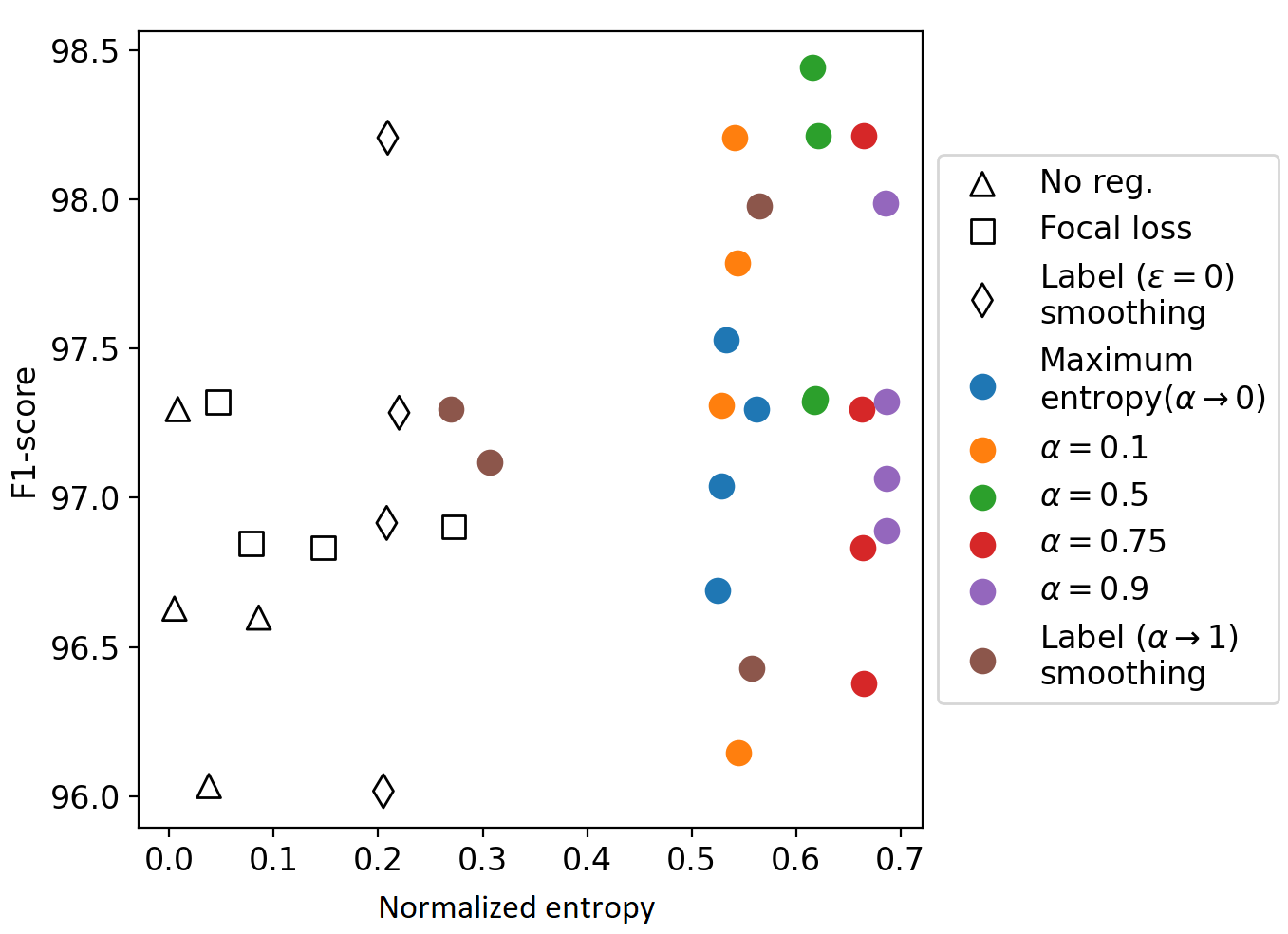}
    \vspace{-.15cm}
    \caption{F1-score as a function of normalized entropy for no regularization, focal loss, maximum entropy learning, $\alpha$-JS, and label smoothing regularization with VGG16, ResNet18, ResNet50, and GoogleNet models.
    }
  \label{fig:H_by_F1}
\end{figure}

In line with this information, one can think that the highest accuracy should be achieved with the model that is closest to the uniform distribution i.e. lower $\alpha$-JS and higher entropy; however this relation is not simple. We observe that higher entropy does not always correspond to higher accuracy. We plotted the F1-score results as a function of normalized entropy and did not observe a monotonic relation between these for ResNet50, VGG16, ResNet18, and GoogleNet architectures as shown in Fig. \ref{fig:H_by_F1}. This complicated relation with $\alpha$ (skewness parameter) and model accuracy motivates us to search for intermediate $\alpha$ values rather than testing only extreme values (label smoothing and maximum entropy regularization).

\section{CONCLUSION}
\label{sec:conclusion}

In this paper, we show that normal vs. defective solders exhibit a low inter-class variance and high intra-class variance; and when compared to the datasets such as ImageNet, they show little diversity. With this information, we tackle the SJI as a fine grained image classification task, and propose $\alpha$-JS based entropy regularization as a solution to FGIC of solder joints. We compare our proposed solution to both the entropy-regularization based approaches as well as the approaches that employ segmentation, attention mechanism, transformer models, and design loss functions for FGIC. We show that the proposed $\alpha$-JS based entropy regularization method achieves the highest and competitive accuracy in SJI across different model architectures. Using activation maps, we show that better class-discriminative regions are obtained with entropy-regularization, and this improves the classification accuracy.



\section*{Appendix}

\subsection{Preliminaries}

\noindent \textbf{Entropy} is a measure of uncertainty of a random variable. 
Entropy of 
a discrete random variable $\mathbf{X}$ is defined by 
Eq. \ref{eq:entropy}, where $\mathbf{x}_{i}$ is the states of $\mathbf{X}$ and $p(\mathbf{X}=\mathbf{x}_{i})= p_{i}$. 

\begin{equation}
    \mathrm{H(p)} = -\sum_{i} p(\mathbf{x}_{i})\mathrm{ln}(p(\mathbf{x}_{i}))
    \label{eq:entropy}
\end{equation}


\noindent \textbf{Cross-entropy} $\mathrm{H}(\mathbf{y},p_{\theta})$ 
between the model distribution $p_{\theta}$ and 
true distribution of the labels $\mathbf{y}$ 
is minimized to train a neural network in supervised classification. $\theta$ is the model parameters. For brevity, cross-entropy for binary class classification is given in Eq. \ref{eq:ce}. 

\begin{equation}
    \mathrm{H}(\mathbf{y}, p_{\theta})= -\mathrm{ln}(p_{t})
    \label{eq:ce}
\end{equation}

where $p_{t}$ is defined as,
\[
    p_{t}= 
\begin{cases}
    p,& \text{if } y_{i}=1\\
    1-p, & \text{otherwise}
\end{cases}
\]

$y_i \in \{0, 1\}$ represents the ground-truth class. $p$ is the probability of the sample to belong to class $y_{i}=1$.  

\noindent \textbf{Kullback-Leibler (KL) divergence} $\mathrm{D_{KL}}(p||q)$ is a difference measure between two probability distributions $p$ and $q$ and defined as in Eq. \ref{eq:KL} for discrete random variables $\mathbf{x}_i$. KL divergence is not symmetric i.e. the forward $\mathrm{D_{KL}}(p||q)$ and reverse KL divergences $\mathrm{D_{KL}}(q||p)$ are not equal.

\begin{equation}
    \mathrm{D_{KL}}(p||q)= \sum_{i}p(\mathbf{x}_i)\mathrm{ln}\left[\frac{p(\mathbf{x}_i)}{q(\mathbf{x}_i)}   \right]
    \label{eq:KL}
\end{equation}

\noindent \textbf{Jensen-Shannon divergence (JSD)} is a symmetric version of KL divergence as given in Eq. \ref{eq:JSD}. 

\begin{equation}
    \mathrm{J}(p||q) = \frac{1}{2}\mathrm{D_{KL}}(p||\frac{p+q}{2}) + \frac{1}{2}\mathrm{D_{KL}}(q||\frac{p+q}{2})
    \label{eq:JSD}
\end{equation}


\subsection{$\alpha$-skew Jensen-Shannon Divergence}
\label{sec:appB}
$\alpha$-JS is obtained by the sum of weighted skewed KL divergences \cite{lee2001effectiveness} of the probability distributions $p$ and $q$ as given in Eq. \ref{eq:skewJSD_full} \cite{nielsen2020generalization}. $\alpha$ 
is a skewness parameter that determines weights to the probability distributions.


\begin{align} \label{eq:skewJSD_full}
    \mathrm{J}_{\alpha}(p||q)
    &= (1-\alpha)\mathrm{D_{KL}}(p||(pq)_\alpha) + \alpha \mathrm{D_{KL}}(q||(pq)_\alpha), \\
    (pq)_\alpha &= (1-\alpha) p+\alpha q  \nonumber \\
    &=-(1-\alpha)\sum p\mathrm{ln} \biggl(\frac{(pq)_\alpha}{p} \biggr) - \alpha\sum q\mathrm{ln} \biggl(\frac{(pq)_\alpha}{q} \biggr)  \nonumber \\
    &= -(1-\alpha)(\sum p\mathrm{ln}(pq)_\alpha - \sum p\mathrm{ln}p) \\& -\alpha(\sum q\mathrm{ln}(pq)_\alpha - \sum q\mathrm{ln}q)  \nonumber \\
    &= -(1-\alpha)\sum p\mathrm{ln}(pq)_\alpha - (1-\alpha)\mathrm{H}(p) \\& - \alpha\sum q\mathrm{ln}(pq)_\alpha -\alpha \mathrm{H}(q)  \nonumber \\
    &= -\sum (1-\alpha)p\mathrm{ln}((1-\alpha)p +\alpha q)  \nonumber \\ &-(1-\alpha) \mathrm{H}(p) - \alpha \sum q\mathrm{ln}((1-\alpha)p + \alpha q) - \alpha \mathrm{H}(q) \nonumber \\
    &= -\sum((1-\alpha)p+\alpha q)\mathrm{ln}((1-\alpha)p + \alpha q) \\& -(1-\alpha) \mathrm{H}(p) - \alpha \mathrm{H}(q)  \nonumber \\
    &= \mathrm{H}((1-\alpha)p + \alpha q) - (1-\alpha) \mathrm{H}(p) - \alpha \mathrm{H}(q) \nonumber
\end{align}

$\alpha$-JS is scaled by $\frac{1}{\alpha (1-\alpha)}$ to guarantee continuity for $\alpha \in  \mathbb{R} \backslash \{0,1\}$ \cite{nielsen2011burbea, nielsen2015total}:

\begin{equation*}
    \mathrm{J}^s_\alpha(p||q)= \frac{1}{\alpha(1-\alpha)} [\mathrm{H}((1-\alpha)p + \alpha q) - (1-\alpha) \mathrm{H}(p) - \alpha \mathrm{H}(q)].
\end{equation*}

\bibliographystyle{IEEEtran}
\bibliography{journal}

\vspace{11pt}

\vspace{-33pt}
\begin{IEEEbiographynophoto}{Furkan Ulger}
received his B.S. and M.S. degree in electrical and electronics engineering from Hacettepe University, Ankara, Turkey, in 2019 and 2022 respectively. 
He is currently working at Aselsan Inc. His research interests include visual inspection, deep learning, image processing and pattern recognition.
\end{IEEEbiographynophoto}

\begin{IEEEbiographynophoto}{Seniha Esen Yuksel} (Senior Member, IEEE) received her B.S. degree in electrical and electronics engineering from the Middle East Technical University, Ankara, Turkey, in 2003; her M.S. degree in electrical and computer engineering from the University of Louisville, USA in 2005; and her Ph.D. degree in computer engineering from the University of Florida, Gainesville, Florida, USA in 2011. Currently, Dr. Yuksel is an associate professor at the Department of Electrical and Electronics Engineering, Hacettepe University, Ankara, Turkey. She is also the director of the Pattern Recognition and Remote Sensing Laboratory (PARRSLAB), where she is doing research on machine learning and computer vision with applications in defense industry. 
\end{IEEEbiographynophoto}

\begin{IEEEbiographynophoto}{Atila Yilmaz} received his MSc degree in Control and Information Technology from Control Systems Centre of UMIST in 1992 and PhD degree from the Biomedical Engineering Division of University of Sussex in 1996. He is currently a Professor with the Department of Electrical and Electronics Engineering of Hacettepe University, Ankara, Turkey. His research interests include artificial intelligence and neural networks in control and biomedical applications especially for image and biomedical signal processing. 
\end{IEEEbiographynophoto}

\begin{IEEEbiographynophoto}{Dincer Gokcen}
(Member, IEEE) received B.S. degree in electrical engineering from Yildiz Technical University in 2005 and Ph.D. degree in electrical engineering from the University of Houston in 2010. He has been with the National Institute of Standards and Technology, Globalfoundries, and Aselsan, before joining Hacettepe University as a Faculty Member in 2016. He also holds an adjunct faculty position at METU MEMS Center. His research interests include nanofabrication, microsystems, and sensors.
\end{IEEEbiographynophoto}

\vfill

\end{document}